\documentclass[letterpaper]{article} 
\usepackage{aaai24}  
\usepackage{times}  
\usepackage{helvet}  
\usepackage{courier}  
\usepackage[hyphens]{url}  
\usepackage{graphicx} 
\usepackage{natbib}  
\usepackage{caption} 
\usepackage{algorithm}
\usepackage{algorithmic}

\usepackage{newfloat}
\usepackage{listings}
\DeclareCaptionStyle{ruled}{labelfont=normalfont,labelsep=colon,strut=off} 
\floatstyle{ruled}
\newfloat{listing}{tb}{lst}{}
\floatname{listing}{Listing}
\title{A Hybrid Classical-Quantum Fine Tuned BERT for
Text Classification}
\author{
   Abu Kaisar Mohammad Masum\textsuperscript{\rm 1},
    Naveed Mahmud\textsuperscript{\rm 2},
   M. Hassan Najafi\textsuperscript{\rm 3},
   Sercan Aygun\textsuperscript{\rm 1} 
}
\affiliations{
    \textsuperscript{\rm 1}School of Computing and Informatics, University of Louisiana at Lafayette, Lafayette, LA, USA\\
    \textsuperscript{\rm 2} College of Engineering and Science, Florida Institute of Technology, Melbourne, FL \\
    \textsuperscript{\rm 3} Electrical, Computer, and Systems Engineering Department, Case Western Reserve University, Cleveland, OH, USA

    c00591145@louisiana.edu, nmahmud@fit.edu, najafi@case.edu, sercan.aygun@louisiana.edu
}

\usepackage{bibentry}
\usepackage{algorithm}
\usepackage{algorithmic}

\usepackage{newfloat}
\usepackage{listings}
\DeclareCaptionStyle{ruled}{labelfont=normalfont,labelsep=colon,strut=off} 
\floatstyle{ruled}
\newfloat{listing}{tb}{lst}{}
\floatname{listing}{Listing}
\usepackage{cite}
\usepackage{amsmath,amssymb,amsfonts}
\usepackage{algorithmic}
\usepackage{booktabs}
\usepackage{tabularx}
\usepackage{graphicx}
\usepackage{textcomp}
\usepackage{xcolor}
\usepackage{multirow}
\usepackage{colortbl}
\usepackage{xcolor}
\usepackage{booktabs}
\usepackage{booktabs}
\usepackage{multirow}
\usepackage{tabularray}

\usepackage{times}
\usepackage{siunitx}
\usepackage{latexsym}
\usepackage{tabularx}
\usepackage{multirow}
\usepackage{booktabs}
\usepackage{pgfplots}
\usepackage{filecontents}
\usepackage{graphicx}
\usepackage{float}
\usepackage{colortbl}
\usepackage[table]{xcolor}
\usepackage{physics}
\usepackage[utf8]{inputenc}
\usepackage{microtype}

\usepackage{xcolor}

\definecolor{darkgreen}{RGB}{0, 100, 0} 

\usepackage{makecell}
\usepackage{array}
\usepackage{multirow}
\usepackage{ragged2e}

\usepackage{rotating}

\usepackage{relsize}

\definecolor{lightgray}{gray}{0.9}
\definecolor{gray}{gray}{0.7}

\usepackage[prependcaption, textsize=footnotesize]{todonotes}

\usepackage{ifthen}
\usepackage{amssymb, stmaryrd, wasysym}
\newboolean{showcomments}
\setboolean{showcomments}{true}
\ifthenelse{\boolean{showcomments}}
{ \newcommand{\mynote}[3]{
     \fbox{\bfseries\sffamily\scriptsize#1}   {\small$\blacktriangleright$\textsf{\emph{\color{#3}{#2}}}$\blacktriangleleft$}}}
{ \newcommand{\mynote}[3]{}}

\usepackage{pifont}
\usepackage{ragged2e}
\usepackage{multirow}
\usepackage{vcell}

\usepackage{textcomp}
\usepackage{tikz}

\usepackage{comment}
\usepackage{hhline}

 \usepackage{url}

\begin{document}

\maketitle

\begin{abstract}
Fine-tuning BERT for text classification can be computationally challenging and requires careful hyper-parameter tuning. Recent studies have highlighted the potential of quantum algorithms to outperform conventional methods in machine learning and text classification tasks. In this work, we propose a hybrid approach that integrates an $n$-qubit quantum circuit with a classical BERT model for text classification. We evaluate the performance of the fine-tuned classical-quantum BERT and demonstrate its feasibility as well as its potential in advancing this research area. Our experimental results show that the proposed hybrid model achieves performance that is competitive with, and in some cases better than, the classical baselines on standard benchmark datasets. Furthermore, our approach demonstrates the adaptability of classical–quantum models for fine-tuning pre-trained models across diverse datasets. Overall, the hybrid model highlights the promise of quantum computing in achieving improved performance for text classification tasks.
\end{abstract}

\section{Introduction}
\label{introduction}

Pre-trained language models such as BERT (Bidirectional Encoder Representations from Transformers) \cite{devlin2018bert}, 
have demonstrated exceptional performance across various language comprehension challenges \cite{sun2019fine}. The BERT model effectively learns universal language representations and can perform Natural Language Processing (NLP) tasks such as text classification even without human supervision \cite{qasim2022fine,garg2020bae,jin2020bert}. However, fine-tuning BERT for specific NLP tasks requires high computational resources and expensive inference time \cite{rogers2021primer, kovaleva2019revealing,zhang2020revisiting}. Usually, fully connected multi-layer perceptron (MLP) networks with many hyper-parameters are required for fine-tuning.

Recently, quantum computing (QC) has been increasingly recognized for its potential impact on conventional machine learning (ML), leading to the development of Quantum Machine Learning (QML). Recent advancements in this field include hybrid classical-quantum methods for transfer learning \cite{mari2020transfer}, text classification \cite{ardeshir2024hybrid}, and speech recognition \cite{qi2022classical}. These studies show how quantum features like \textit{superposition} and \textit{entanglement} can help solve various ML problems.

In this work, we propose a hybrid classical-quantum approach for classifying texts. We use an $n$-qubit quantum circuit to fine-tune a classical BERT model. 
We perform a comprehensive evaluation of the proposed method for text classification on a variety of popular datasets. The BERT model was selected for its outstanding performance in understanding the language context bidirectionally. It was pre-trained with vast text data, making it adaptable for tasks with minimum data labels. BERT's flexibility, availability, and transfer learning capabilities suit our proposed hybrid classical-quantum method. The proposed approach entails using pre-trained BERT embeddings and combining them with quantum techniques to improve the overall performance of text classification tasks. This approach leverages classical NLP and QC to improve accuracy and efficiency in handling complex text information. 

While current quantum hardware is not yet fault-tolerant enough to achieve actual computational speedup, we perform experiments with quantum circuit simulations to validate the proposed hybrid approach. These experiments yield classification accuracy scores equal to or better than those achieved by conventional fine-tuning techniques. This research contributes to several key areas: \\
- Integrating quantum circuits with classical BERT models for text classification \\
- Evaluating performance across various quantum circuit configurations \\
- Highlighting the potential of leveraging QC in language models for future advancements.

This article is organized as follows: Section II gives an overview of QC 
Fundamentals, Quantum Gates, Variational Quantum Circuits, and Quantum ML Literature. Section III presents the proposed approach for hybrid classical-quantum text classification. Section IV describes the experimental setup, including the datasets used, model training procedures, and key parameters. It further discusses experimental results, comparing the performance of the proposed model against classical counterparts across various datasets. Finally, Section V summarizes our findings, highlights limitations, and suggests directions for future work in the intersection of QML.

\section{Background}
\label{one}

\subsection{Quantum Computing Fundamentals}
Two distinctive features of QC are \textit{superposition} and \textit{entanglement} \cite{preskill2018quantum,jozsa1997entanglement}. A qubit is characterized by the Boolean states $0$ and $1$ within a quantum system, represented by a normalized quantum state $\left|0\right\rangle$ and $\left|1 \right\rangle$ \cite{schumacher1995quantum}. In computational terms, any state formed by the combination of these two states is referred to as a superposition denoted by $\alpha\left|0\right\rangle$ + $\beta\left|1\right\rangle$, where $\alpha$ and $\beta$ satisfy the condition $\left| \alpha \right|^2 + \left| \beta \right|^2 = 1$. However, entanglement is a special correlation between qubits where the state of one qubit (e.g., \textit{qubit A}) instantly affects another (e.g., \textit{qubit B}), in any case of distance. The combined wave function representing entangled particles is denoted by $\Psi_{AB} = (1/\sqrt{2})(\Psi_{1A} \Psi_{2B} - \Psi_{2A} \Psi_{1B})$.

\subsection{Quantum Gates}
Quantum gates are the fundamental components of
quantum circuits are like classical logic gates in classical computing. They operate on qubits and manipulate their quantum states by leveraging properties like superposition and entanglement. A quantum circuit is a sequence of quantum gates \cite{kais2014introduction}. These gates are used for data encoding from classical to quantum; for example, the Hadamard gate ($H$ gate) creates superposition states. A rotation gate like $R_Y(\theta)$ gate is a parameterized gate that rotates the qubit along the y-direction by the specified angular parameter $\theta$. Two-qubit gates like the \texttt{CNOT} gate have a control qubit and a target qubit. The \texttt{NOT} (inversion) operation is implemented on the target qubit based on the value of the control qubit. The \texttt{CNOT} is particularly useful for creating an entanglement effect between qubits \cite{chen2020hybrid}.

\subsection{Variational Quantum Circuits}
\begin{figure*}[h]
    \centering
\includegraphics[width=\linewidth]{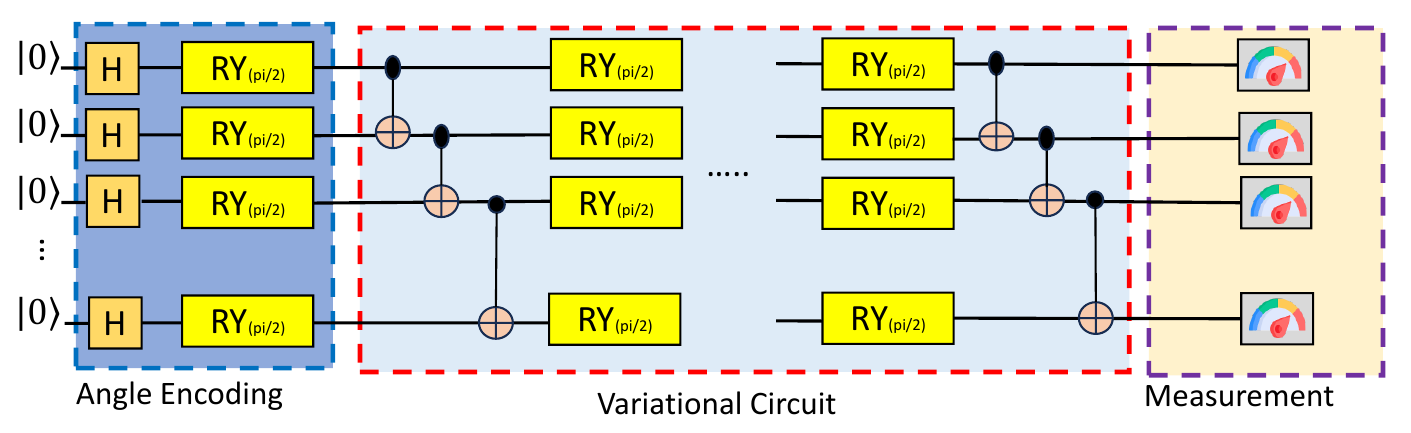}
    \caption{Angle encoding, variational circuit, and the measurement steps in an example VQC.}
    \label{fig:enter-label}
    \vspace{-10pt}
\end{figure*}

A Variational Quantum Circuit (VQC) allows for the exploration of different quantum states and state transformations by varying the parameters of its gates. Measurements extract classical information from the final quantum state and are then used to evaluate a cost function. The parameters are then varied or \emph{trained} to minimize the cost function. Hence, parameterized VQCs can be trained for ML tasks \cite{saxena2022performance, rizvi2023neural}, analogous to the training of neural networks.
VQCs have a relatively smaller number of parameters compared to classical neural networks, making them computationally efficient and easier to optimize \cite{chen2022variational, ajagekar2024variational}. In Figure \ref{fig:enter-label}, a VQC example is illustrated.

\subsection{Quantum ML Literature}
QC and its applications in ML span areas such as computer vision and language processing. In recent years, challenges in document classification have led to the growth of conventional neural network-based models alongside the rise of quantum ML systems. Techniques like BERT stand out within the established world of pre-trained models, proving their usefulness in achieving cutting-edge performance across varied tasks such as NLP and natural language generation models \cite{devlin2018bert,adhikari2019docbert,qu2019bert,zhang2019bertscore,souza2019portuguese}. On the other hand, the rapid advancement of QC has revealed different aspects demonstrating quantum benefits in quantum ML \cite{schuld2015introduction}. These include more complex feature representation and improved security for ML model parameters. This advancement is based on the use of variational quantum circuits, which have demonstrated the competitive performance of text classification \cite{yang2022bert}. 

In addition, several ML tasks, such as Omar et al. \cite{omar2023quantum}, have used quantum ML to classify Arabic text data, Quantum self-attention \cite{li2024quantum,zhang2024light} for text classification, and hybrid transfer learning \cite{ardeshir2024hybrid}. Li et al.\cite{li2022quantum} recently proposed an attention-based quantum natural language processing (QNLP) technique. Other tasks of QNLP include ensemble learning by Bouakba et al. \cite{bouakba2023ensemble}, Quantum temporal convolution by Yang et al. \cite{yang2022bert}, the exploration of quantum algorithms by Zeng et al. \cite{zeng2016quantum}, and Lorenz et al. \cite{lorenz2023qnlp}.

Quantum-inspired ML has emerged as a promising direction for enhancing 
classical ML models~\cite{felser2021quantum}. Achieving optimal performance in these approaches requires 
smooth input preparation to achieve the best performance~\cite{zhao2021smooth}. For example, quantum-inspired methods have been applied to cancer classification with encouraging results~\cite{pomarico2021proposal}. 
Recently, in 6G network research, quantum-inspired models have demonstrated superior capabilities in data processing and training~\cite{duong2022quantum}. 
A wide range of applications, including optimization, binary classification,  protein design, and clonogenic assay evaluation~\cite{li2017quantum, tiwari2018towards, panizza2024protein, sergioli2021quantum}, now leverage quantum principles to enhance computational efficiency and accuracy, offering more effective solutions to complex problems.

\section{Proposed Approach}

\begin{figure*}[t]

  \includegraphics[width=\linewidth]{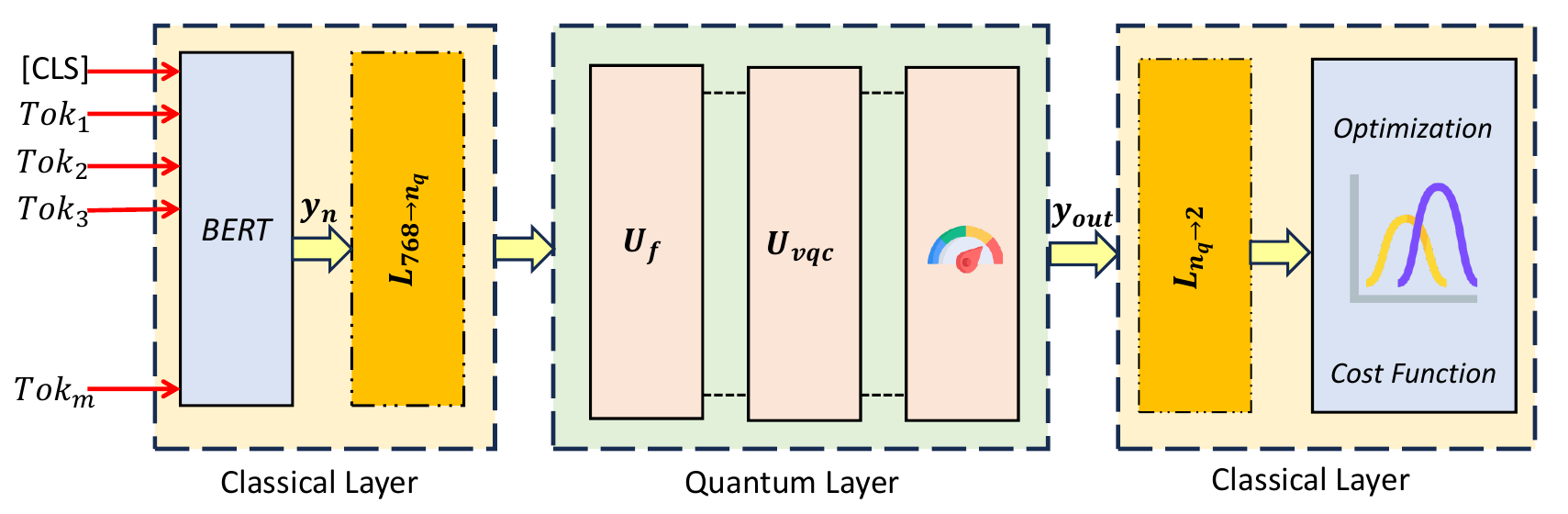}
  \caption{Overview of our approach for a hybrid classical-quantum BERT model.}
  \label{overview}
  \vspace{-10pt}
\end{figure*}

In this work, we target a text classification problem using quantum ML. Our proposed approach for text classification incorporates the classical BERT model and quantum layer. The big picture of the overall proposal is shown in Figure \ref{overview}. The goal 
is to leverage the power of QC to enhance the performance of traditional text classification tasks. The architecture consists of three main layers:
\begin{itemize}
    \item \textbf{Input Classical Layer:} The first layer processes the input text data using a pre-trained language model like BERT. The output of this layer is a sequence of embeddings \cite{feng2020language} representing the semantic meaning of each word in the text.
    \item \textbf{Quantum Layer:} The second layer is where the quantum computations take place. This layer includes a feature encoding circuit \cite{schuld2019quantum} with a quantum variational circuit. The feature encoding circuit maps the classical text embeddings from the previous layer into a quantum state. The quantum variational circuit then applies a series of parameterized quantum gates to manipulate this state.
    \item \textbf{Output Classical Layer:} This final layer measures the quantum state to fine-tune resulting measurement outcomes into a classification probability distribution. The most likely class is chosen as the predicted label for the input text.
\end{itemize}

\subsection{Classical BERT Model}

In the classical BERT, a fully connected layer links the last hidden state of the model (first output) to the model's input tokens.  
The output ${y}_{n}$ is produced by applying a non-linear activation function $f$ to the weighted sum of the input ${x}_{n}$, leveraging a set of ${\theta}_{n}$ parameters, corresponding to each layer's weights and biases.
\begin{equation}
    {L}_{{n}_{0}\rightarrow {n}_{1}}:{y}_{n} = f({x}_{n},{\theta}_{n})
\end{equation}
where, ${L}_{{n}_{0}\rightarrow {n}_{1}}$ indicates a change in dimensions from ${n}_{0}$ to ${n}_{1}$.
The BERT model we used returns a $n_1=768$ dimensional vector. For fine-tuning the model output for classifying texts, we integrate a quantum layer with the BERT output. 

\subsection{Fine-tuning with Quantum Layer}
In our approach, we introduce an additional pooling layer, ${L}_{{n}_{1}\rightarrow n_q}$, prior to the quantum layer to reduce the feature dimensionality of the BERT output from 768 to $n_q$. Each data feature $d_i$ in the resulting layer is encoded and represented by a single qubit through the following transformation:

\begin{equation}
    U_{f}\left|0 \right\rangle = \bigotimes_{i=1}^{n_q} \cos(d_i)\ket{0} + \sin(d_i)\ket{1}
\end{equation}

\noindent where $U_{f}$ represents the quantum feature encoding circuit (see Figure \ref{overview}). $U_{f}$ is realized using $R_y$ rotation gates with feature data $d_i$ from the previous classical layer used as the rotation angles.  
The resulting state is denoted as $\ket{x}$.
We apply a shallow depth, variational quantum circuit \cite{cerezo2021variational,yang2022bert} with parameter set $\theta_{n}$  that maps $\left|x \right\rangle$ to $\left|y \right\rangle$.
\begin{equation}
   {\mathcal{L}}:\left|x \right\rangle\rightarrow \left|y \right\rangle = U_{vqc}\left|x \right\rangle
\end{equation}
where $U_{vqc}$ represents the VQC consisting of trainable parameters. The structure of the VQC circuit is shown in Figure \ref{overview}. The first layer of the circuit consists of $H$ gates that create a superposition state, allowing the circuit to sample multiple solutions in parallel for the next layer, which contains the trainable parameterized rotation gates. This speeds up the optimization process of the variational parameters during the training phase. For parameterized rotation, we use $R_y$ gates that affect the real amplitudes of the quantum state.   
The next layer of \texttt{CNOT} gates incorporates entanglement in the circuit, amplifying the model's ability to capture complex correlations in the encoded data. 
In a fully entangled state, all qubits in the system are maximally correlated with each other.
Now, to extract classical output ${y}_{out}$ as a vector from the quantum circuit, we measure the expectation values of qubits. We define this as a measurement layer of the quantum circuit, thus converting the final quantum state into a classical vector.
\begin{equation}
    M:\left|{y} \right\rangle\rightarrow {y}_{out} = \left\langle \widehat{O} \right\rangle{y}
\end{equation}
Here $\left\langle \widehat{O} \right\rangle{y}$ represents the expectation value of an observable operator $\left\langle \widehat{O} \right\rangle$ in the state $\left|{\psi}_{out} \right\rangle$.
After measurement, we apply ${{L}_{n}}_{q\rightarrow 2}$, a classical pooling layer that transforms ${n}$ features into two features that are used in the final prediction. 
The final prediction is compared with data labels to evaluate a mean-square-error loss function. Based on the loss, we calculate quantum gradients to optimize the VQC parameters to minimize the loss.

\section{Experiment Setup and Results Discussion}
\label{experiment}

\subsection{Datasets} We collected five different datasets from publicly available sources for text classification: $5000$ IMDb data from \linebreak \url{https://huggingface.co/datasets/imdb}, $5572$ email Spam data from \url{https://www.surgehq.ai/datasets/email-spam-dataset} , $6000$ SST from \url{https://nlp.stanford.edu/sentiment/treebank.html}, $10,000$ Yelp data from \url{https://www.yelp.com/dataset}, and $11,541$ Twitter data from \url{https://huggingface.co/datasets/osanseviero/twitter-airline-sentiment}. 

\subsection{Model Training} For classical-quantum BERT training, we leverage the NVIDIA T4 GPU. The subsequent experiments are run using PyTorch 2.1.0+cu118, which serves as the foundational framework. We use transformer version 4.35.2 for BERT, and Scikit-learn 1.2.2 for model selection and classification report analysis. In the quantum part, we use pennylane 0.33.1 with the default simulator to create and run quantum circuits.

\subsection{Model Parameters} 
Table~\ref{tab:quantum-parameters} outlines the crucial classical and quantum parameters employed in our hybrid BERT model settings. These carefully chosen settings were instrumental in achieving optimal performance across diverse datasets. 

\begin{table*}[h]
  \centering
  \begin{tabular}{ll|ll}
    \toprule
    \multicolumn{2}{c|}{\textbf{Quantum Settings}} & \multicolumn{2}{c}{\textbf{Training Settings}} \\
    \textbf{Parameter} & \textbf{Value} & \textbf{Parameter} & \textbf{Value} \\
    \midrule
     Qubits Range         & 2--10                & Epoch                & 10 \\
    Circuit Depth        & 4                    & Batch Size           & 32 \\
    Quantum Weight       & 0.01                 & Max Sequence Length  & 25 \\
    Quantum Gates        & H, Y-rotation, CNOT  & Learning Rate        & $1 \times 10^{-3}$ \\
    Encoding Method      & Angle Encoding       & \multicolumn{2}{c}{} \\
    Entanglement         & Yes                  & \multicolumn{2}{c}{} \\
    Feature Map          & Pauli-Z              & \multicolumn{2}{c}{} \\
    Simulator            & Default              & \multicolumn{2}{c}{} \\
    \bottomrule
  \end{tabular}
   \caption{Key Parameters for Hybrid Classical-Quantum BERT.}
 \label{tab:quantum-parameters}
\end{table*}

Our classical model includes 10 epochs, a batch size of 32,  a maximum sequence length of 25, and a learning rate of \(1 \times 10^{-3}\). These parameters balance training speed, stability, and model generalization. For our experiments, we used 2 to 10 qubits with a circuit depth of 4, where we leveraged H gates, Y-rotation gates, and \texttt{CNOT} gates to manipulate the qubits effectively. Moreover, we employed angle encoding for data representation, while quantum entanglement was used to capture complex correlations within the data. A quantum weight of 0.01 is used to integrate quantum features without overwhelming the classical model. Again, the Pauli-Z feature map is used to enhance the model's ability to learn nonlinear relationships, and a default quantum simulator ensures accurate simulations of our model. In summary, these parameters create a robust hybrid architecture that integrates the strengths of classical and QC to ensure accurate model performance.

\subsection{Results Comparison} 
In the experiments, we compare the proposed classical-quantum BERT to a simple classical BERT (baseline), and a Multilayer Perceptron (MLP) fine-tuned BERT across IMDB, Spam, SST, Yelp, and Twitter datasets. An in-depth analysis across the datasets reveals that the proposed classical-quantum BERT consistently performs better than or equal to its counterparts. The table summary focuses on the specific strengths and weaknesses of the implemented models in text classification tasks.

\begin{table*}[h]
\centering

\begin{tabular}{l|cc|cc|cc}
\toprule
\textbf{Dataset} & \multicolumn{2}{c|}{\textbf{BERT\textsubscript{Classical-Quantum}}} & \multicolumn{2}{c|}{\textbf{BERT\textsubscript{Classical}}} & \multicolumn{2}{c}{\textbf{MLP\textsubscript{tf-idf}}} \\
\cmidrule{2-7}
& \textbf{Train F1} & \textbf{Val. F1} & \textbf{Train F1} & \textbf{Val. F1} & \textbf{Train F1} & \textbf{Val. F1} \\
\midrule
IMDb & 0.64 & 0.64 & 0.63 & 0.63 & 0.63 & 0.63 \\
Spam & 0.95 & 0.95 & 0.95 & 0.95 & 0.95 & 0.95 \\
Twitter & 0.86 & 0.86 & 0.83 & 0.83 & 0.82 & 0.83 \\
SST & 0.82 & 0.82 & 0.82 & 0.82 & 0.70 & 0.72 \\
Yelp & 0.79 & 0.79 & 0.78 & 0.78 & 0.78 & 0.75 \\
\bottomrule
\end{tabular}
\caption{Comparison of performance metrics (Train F1 and Validation F1) for three different models (BERT\textsubscript{Classical-Quantum}, BERT\textsubscript{Classical}, and MLP\textsubscript{tf-idf}) across multiple datasets (IMDb, Spam, Twitter, SST, and Yelp). The table presents the training and validation F1 scores for each model and dataset. The values in the table represent the F1 scores, with important scores highlighted. Changes in performance can be observed across the models and datasets. BERT\textsubscript{Classical-Quantum} represents the proposed quantum-enhanced version of the BERT model, while BERT\textsubscript{Classical} denotes the classical BERT model. MLP\textsubscript{tf-idf} refers to a Multilayer Perceptron model using tf-idf features.}
\label{tab:scores_}
 \vspace{-15pt}
\end{table*}

As shown in Table~\ref{tab:scores_}, our classical-quantum BERT demonstrates consistently high performance, particularly on datasets like IMDb, Spam, and Twitter. It achieves an F1 score of 0.64 for both training and validation on the IMDb dataset, which is marginally better than both the classical BERT and MLP models, scoring 0.63 in each. On the Spam dataset, all three models achieve high F1 scores of 0.95 for both training and validation, indicating that this task is better for each model’s architecture. However, the proposed BERT outperforms its counterparts for the Twitter dataset, getting an F1 score of 0.86 for both training and validation compared to the classical BERT’s score of 0.83. The MLP’s score is 0.82 F1 score for training and 0.83 for validation on the same dataset.

In contrast, on the SST dataset, both the classical-quantum BERT and classical BERT achieve the same F1 scores of 0.82 for both training and validation. The MLP model for the dataset underperforms, with scores of 0.70 for training and 0.72 for validation F1 score. This result indicates that while the quantum-enhanced model does not always exceed the performance of its classical counterpart, it still remains competitive. For the Yelp dataset, the classical-quantum BERT shows a slight advantage. It scores 0.79 for both training and validation F1 score, compared to the classical BERT’s score of 0.78 and the MLP’s 0.78 for training, as well as 0.75 for validation.

\subsection{Comparative Analysis} We compare the accuracy of the proposed classical-quantum BERT model across various datasets, qubit values, and evaluation with existing models (see Table \ref{tab:my-table}). This table suggests the potential of the proposed model in achieving high accuracy for various text classification tasks with the flexibility of adjusting qubit values. 
\begin{table*}[h]
\centering
\setlength{\tabcolsep}{4pt}  
\begin{tabular}{lccc}
\hline
\textbf{Method} & \textbf{Qubits} & \textbf{Dataset} & \textbf{Accuracy} \\ \hline
Lambeq \cite{kartsaklis2021lambeq} & 3 & Spam & 0.93 \\ 
& & SST & 0.59 \\ \hline
Hybrid QBERT \cite{ardeshir2024hybrid} & 5 & Spam & 0.97 \\ \hline
Adapting QBERT \cite{li2023adapting} & 10 & SST & 0.89 \\ \hline
\multirow{5}{*}{\textbf{Proposed}} & \multirow{5}{*}{2-10} & IMDB & \cellcolor[gray]{0.9}0.67 $\pm$ 0.04 \\ 
& & Spam & \cellcolor[gray]{0.9}0.96 $\pm$ 0.03 \\ 
& & Twitter & \cellcolor[gray]{0.9}0.86 $\pm$ 0.02 \\ 
& & SST & \cellcolor[gray]{0.9}0.82 $\pm$ 0.02 \\ 
& & Yelp & \cellcolor[gray]{0.9}0.79 $\pm$ 0.01 \\ \hline
\end{tabular}
\caption{Performance comparison of proposed model with related work for varying qubits and diverse datasets. Includes Lambeq \cite{kartsaklis2021lambeq}, Hybrid QBERT \cite{ardeshir2024hybrid}, Adapting QBERT \cite{li2023adapting}, and Proposed (2-10). Reports accuracy for Spam, SST, IMDB, Twitter, and Yelp. For the proposed model, mean accuracy with standard deviations (±) is provided and exhibits comparable performance across data sets with respect to related work.}
\label{tab:my-table}
\vspace{-10pt}
\end{table*}

The Lambeq model \cite{kartsaklis2021lambeq} using 3 qubits achieves an accuracy of 0.93 on the Spam dataset but only 0.59 on the SST dataset. The hybrid QBERT \cite{adhikari2019docbert} that uses 5 qubits achieves a higher accuracy of 0.97 on the Spam dataset, showing the effectiveness of increasing the complexity of the quantum circuit. Adapting QBERT \cite{li2023adapting} with 10 qubits shows a significant improvement in the SST dataset with an accuracy of 0.89, indicating the benefits of a quantum model for more text datasets.

Our proposed classical-quantum BERT model demonstrates flexibility by using a range of 2 to 10 qubits and consistently performs well across all datasets. Specifically, it achieves a mean accuracy of 0.67 ± 0.04 on IMDB, 0.96 ± 0.03 on Spam, 0.86 ± 0.02 on Twitter, 0.82 ± 0.02 on SST, and 0.79 ± 0.01 on Yelp. These results suggest that our model maintains a competitive edge, either outperforming or matching the accuracy of existing models across all tested datasets. 

\subsection{Training Time} 
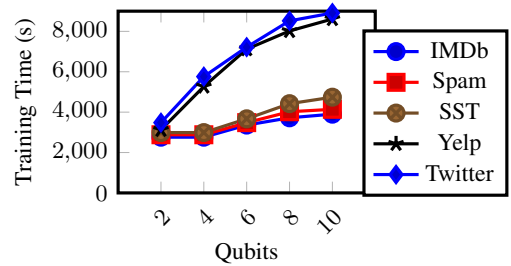
\begin{figure}[!h]
  \centering
  \begin{tikzpicture}
    \begin{axis}[
      xlabel={Qubits},
      ylabel={Training Time (s)},
      legend pos=north west,  
      grid style={dashed,gray!40},  
      width=5cm,               
      height=4cm,               
      tick label style={font=\small},
      label style={font=\small},
      legend style={font=\small, at={(0.95,0.90)}},  
      xmin=0, xmax=12,
      ymin=0, ymax=9000,
      xtick={2,4,6,8,10},
      ytick={0,2000,4000,6000,8000,10000},
      y tick label style={
        /pgf/number format/.cd,
        fixed,
        fixed zerofill,
        precision=0,
        /tikz/.cd
      },
      line width=1.2pt,            
      mark size=3pt,               
      xticklabel style={rotate=45, anchor=north east},  
    ]
    \addplot table {
      2 2760
      4 2760
      6 3360
      8 3720
      10 3900
    };
    \addlegendentry{IMDb}

    \addplot table {
      2 2880
      4 2880
      6 3480
      8 4020
      10 4140
    };
    \addlegendentry{Spam}

    \addplot table {
      2 2988
      4 2988
      6 3670
      8 4420
      10 4740
    };
    \addlegendentry{SST}

    \addplot table {
      2 3126
      4 5269
      6 7120
      8 8020
      10 8612
    };
    \addlegendentry{Yelp}

    \addplot table {
      2 3480
      4 5760
      6 7220
      8 8520
      10 8910
    };
    \addlegendentry{Twitter}
    \end{axis}
  \end{tikzpicture}
  \caption{The graph shows training time variations in a classical-QC model across datasets (IMDb, Spam, SST, Yelp, Twitter) with different qubit values. IMDb, Spam, and SST maintain stable times, while Yelp and Twitter see increased times with more qubits. Key values: IMDb (2 qubits: 2760s, 10 qubits: 3900s), Spam (2 qubits: 2880s, 10 qubits: 4140s), SST (2 qubits: 2988s, 10 qubits: 4740s), Yelp (2 qubits: 3126s, 10 qubits: 8612s), Twitter (2 qubits: 3480s, 10 qubits: 8910s). Insights into qubit count and training time relationships reveal dataset-specific differences in the classical-QC model.}
  \label{fig:linegraph}
  \vspace{-20pt}
\end{figure}

We estimate the training time for the hybrid models as depicted in the line graph in Figure \ref{fig:linegraph}. This graph provides insights into the training time evolution of a classical-QC model over a variety of datasets with varying qubit values. The x-axis shows the number of qubits used in the classical-quantum model, while the y-axis represents the training duration in seconds. The consistent training times suggest that the model's performance is relatively stable and insensitive to changes in the number of qubits.
IMDb dataset shows a training time of 2760 seconds with 2 qubits, which slightly increases to 3900 seconds with 10 qubits. Similarly, Spam and SST datasets display modest increases in training times, indicating minimal sensitivity to the number of qubits used. The graph reveals a notable increase in training times for the Yelp and Twitter datasets as the number of qubits rises. For Yelp data, training time starts at 3,126 seconds with 2 qubits and escalates significantly to 8,612 seconds with 10 qubits. A similar trend is observed for the Twitter dataset, where the training time jumps from 3,480 seconds at 2 qubits to 8,910 seconds at 10 qubits. This suggests that these datasets may involve more complex patterns or higher data dimensionality, requiring longer training times as quantum circuit complexity increases.

\section{Conclusion \& Limitations}
\label{conclusion}

This work evaluates the interaction between pre-trained language models and quantum computing in Quantum Machine Learning (QML). We combined a pre-trained BERT model with a quantum variational circuit for text classification. Despite scalability, computing resource constraints, and longer training time, our study highlights the importance of addressing potential quantum advantage over classical models and achieves acceptable accuracy results across a variety of datasets. The results indicate that while the hybrid model demonstrates promising performance, it also reveals areas where further optimization could be beneficial, particularly in improving the computational efficiency and reducing the training time. The hybrid classical-quantum BERT model has limits besides these positive results. Scalability is limited by quantum devices, and the integrated circuit takes substantially longer to train than classical models. However, this approach is an important step in leveraging classical-quantum collaboration for improved document classification, paving the way for higher efficiency when dealing with complicated textual content.

\section{Acknowledgments}
This work is supported by the National Science Foundation under Grant No. 2019511 and generous gifts from NVIDIA.


\bibliography{aaai24}

\end{document}